\documentclass[lettersize,journal]{IEEEtran}
\usepackage{amsmath,amsfonts}
\usepackage{algorithmic}
\usepackage{algorithm}
\usepackage{array}
\usepackage[caption=false,font=normalsize,labelfont=sf,textfont=sf]{subfig}
\usepackage{textcomp}
\usepackage{stfloats}
\usepackage{url}
\usepackage{verbatim}
\usepackage{graphicx}
\usepackage{subcaption}
\usepackage{caption}
\usepackage{float} 
\usepackage{xcolor}
\usepackage{multirow}
\usepackage{multicol}
\usepackage{hyperref}
\usepackage{cite}
\newcommand{\etal}{\textit{et al.}}
\begin{document}

\title{Learning to Tune Like an Expert: Interpretable and Scene-Aware Navigation via MLLM Reasoning and CVAE-Based Adaptation}

\author{Yanbo Wang, Zipeng Fang, Lei Zhao, Weidong Chen
        % <-this % stops a space
% \thanks{This paper was produced by the IEEE Publication Technology Group. They are in Piscataway, NJ.}
}

% The paper headers
\markboth{Journal of \LaTeX\ Class Files,~Vol.~14, No.~8, August~2021}%
{Shell \MakeLowercase{\textit{et al.}}: A Sample Article Using IEEEtran.cls for IEEE Journals}

% \IEEEpubid{0000--0000/00\$00.00~\copyright~2021 IEEE}
% Remember, if you use this you must call \IEEEpubidadjcol in the second
% column for its text to clear the IEEEpubid mark.

\maketitle

\begin{abstract}
Service robots are increasingly deployed in diverse and dynamic environments, where both physical layouts and social contexts change over time and across locations. In these unstructured settings, conventional navigation systems that rely on fixed parameters often fail to generalize across scenarios, resulting in degraded performance and reduced social acceptance. Although recent approaches have leveraged reinforcement learning to enhance traditional planners, these methods often fail in real-world deployments due to poor generalization and limited simulation diversity, which hampers effective sim-to-real transfer. To tackle these issues, we present LE-Nav, an interpretable and scene-aware navigation framework that leverages multi-modal large language model reasoning and conditional variational autoencoders to adaptively tune planner hyperparameters. To achieve zero-shot scene understanding, we utilize one-shot exemplars and chain-of-thought prompting strategies. Additionally, a conditional variational autoencoder captures the mapping between natural language instructions and navigation hyperparameters, enabling expert-level tuning. Experiments show that LE-Nav can generate hyperparameters achieving human-level tuning across diverse planners and scenarios. Real-world navigation trials and a user study on a smart wheelchair platform demonstrate that it outperforms state-of-the-art methods on quantitative metrics such as success rate, efficiency, safety, and comfort, while receiving higher subjective scores for perceived safety and social acceptance. Code is available at \href{https://github.com/Cavendish518/LE-Nav}{https://github.com/Cavendish518/LE-Nav}.

Note to Practitioners—Service robots often experience degraded performance of traditional local planners due to changing and dynamic environmental conditions during navigation. This work investigates automatic hyperparameter tuning for planners such as DWA and TEB, and our framework LE-Nav can be used to adjust hyperparameters of any optimization-based planner. Existing navigation frameworks are typically either end-to-end, lacking safety guarantees, or rely on reinforcement learning-based tuning with limited generalization. Our method integrates conditional generative modeling with multi-modal large language model (MLLM) reasoning. By designing two prompting strategies, we enable the MLLM to generate stable and accurate scene descriptions. We use a conditional variational autoencoder to learn human expert tuning strategies, enhanced with data augmentation and attention masking to address inevitable MLLM packet loss in real applications. The decoupling of the MLLM and action modules improves decision transparency, allowing clear insight into how scene analysis informs navigation behavior. Experiments demonstrate that our method adaptively generates hyperparameters comparable to human experts, while being robust to packet loss and compatible with various MLLMs. Navigation trials and user studies further validate significant improvements over existing methods in safety, efficiency, comfort, robustness, transparency, and user satisfaction. Future work includes enhancing real-time scene understanding with advanced MLLMs, expanding support to more planners with personalized tuning, and extending the framework to collaborative multi-robot systems.
\end{abstract}

\begin{IEEEkeywords}
Interpretable navigation, adaptive learning, large language models, decision transparency, service robots.
\end{IEEEkeywords}

\section{Introduction}
In recent years, service robots~\cite{servicerob,wang2024long,deng2025neslam} have been increasingly deployed in dynamic and unstructured human-centric environments such as hospitals, shopping malls, and hotels. Unlike industrial settings where the operating conditions are relatively fixed, service robots must navigate in scenarios that are constantly changing due to varying layouts and human activities~\cite{wang2024long}. In such environments, the abilities to adapt to situation-specific navigation and effectively manage human-robot interactions are essential for ensuring robust performance and socially acceptable behavior.

However, most existing industrially mature navigation systems employ planners with fixed parameters that require manual tuning for specific environments. These systems often struggle to generalize across diverse and dynamic situations. This limitation not only compromises navigation performance but also undermines user trust and social acceptance. 

To address these issues, previous researchers have explored two main approaches. One approach~\cite{hu2023planning,zhengdiffusion,drltase} involves end-to-end navigation frameworks, which significantly improve adaptability to complex environments. However, such methods often lack formal guarantees for safety and decision transparency. The other approach~\cite{dobrevski2020adaptive,dadwa,patel2021dwa} integrates data-driven techniques with traditional planning algorithms, aiming to combine the adaptability of learning-based methods with the reliability and safety assurances of classical planners.

In this work, we adopt the second approach, which focuses on integrating learning-based techniques with traditional optimization-based planners. Most existing frameworks~\cite{dobrevski2020adaptive,dadwa,patel2021dwa} following this direction rely on reinforcement learning to adjust the hyperparameters of optimization-based planners. While these methods have shown promising results in controlled environments, our empirical observations during real-world deployment indicate that they still suffer from two major limitations: (1) the sim-to-real gap, which hinders direct deployment in real-world scenarios, and (2) limited simulation diversity, which restricts the generalization ability of the learned models.

\begin{figure*}[t]
  \centering
\includegraphics[width=0.9\linewidth]{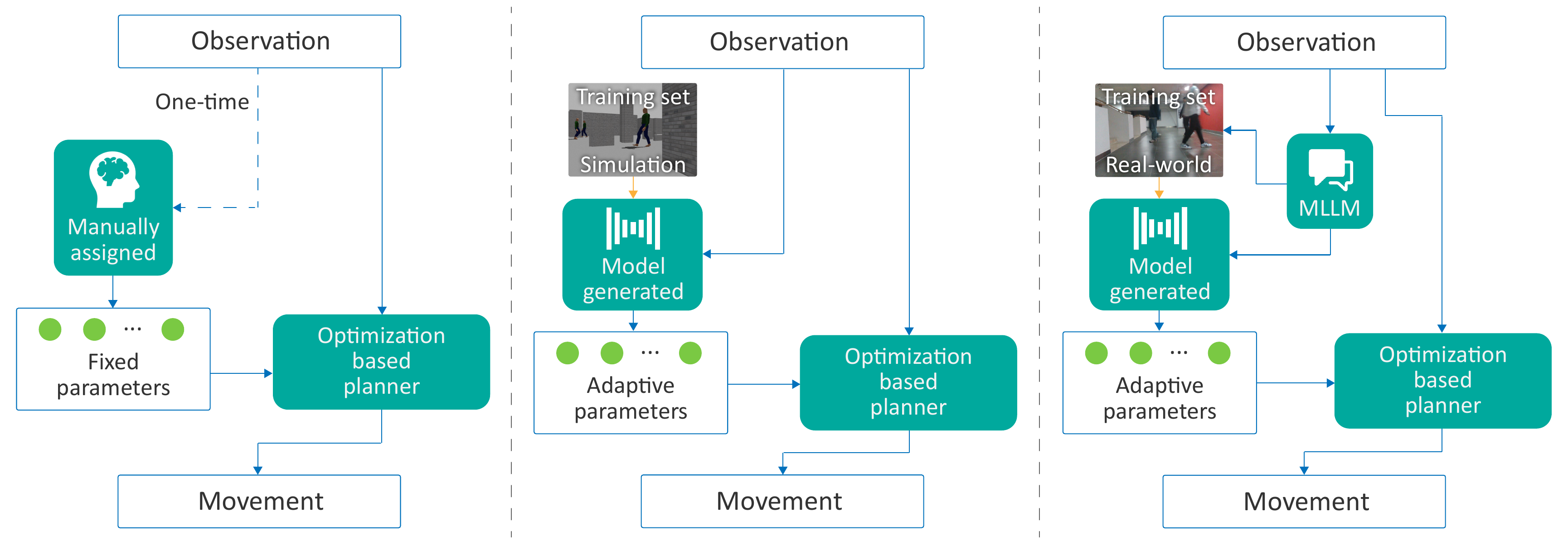}
        \caption{Comparison of the traditional optimization-based planner, the reinforcement learning-enhanced optimization-based planner, and our proposed framework. The CVAE-based generative model can be directly trained on real-world datasets, thereby avoiding concerns about the sim-to-real gap. In addition, the use of MLLM enhances the generalization capability of the framework beyond the limitations of predefined simulation scenarios.}
        \label{fig1:compare}
\end{figure*}

These observations motivate us to develop a novel scene-aware navigation framework that adaptively adjusts the hyperparameters of an optimization-based planner, with generalizability for service robots in real-world environments. To enhance the framework's generalization capability, we first incorporate a multi-modal large language model (MLLM) to perform scene reasoning based on both visual observations and predefined domain-specific rules. We then formulate the hyperparameter adjustment task as a multi-modal conditional generation problem, in which the MLLM-derived scene description serves as the condition for a conditional variational autoencoder (CVAE) to generate suitable hyperparameters for  planners. Notably, since service robots frequently interact with humans, our combination of MLLM-based semantic reasoning and CVAE-based learning from expert data offers strong interpretability of the resulting navigation behaviors. We compare our framework with previous approaches in Fig.~\ref{fig1:compare}.

We collected hyperparameters that human experts adjusted for different planners in various real-world scenarios. After learning these expert strategies, our method could generate planner-specific hyperparameters in new scenarios. Evidence shows that the generated hyperparameters not only match the performance level of human experts, but also exhibit greater continuity and smoothness. We further conducted over a hundred real-world navigation trials, comparing our approach against state-of-the-art (SOTA) methods. Quantitative evaluations across multiple metrics, including success rate, efficiency, safety, and comfort, demonstrate that our method achieves superior overall navigation performance. Notably, it generalizes well to unseen environments, and exhibits strong adaptability in long-horizon tasks. In addition, we have performed a user study using a smart wheelchair platform. Subjective evaluations from both pedestrians and wheelchair users indicate that our method achieves higher scores in terms of perceived safety and social acceptance, further validating its real-world applicability and human-centered design.

We summarize our contributions as below:
\begin{itemize}
    \item {We propose a novel adaptive navigation framework LE-Nav for service robots that supports both environment-aware and user-aware autonomous hyperparameter tuning. LE-Nav preserves the trajectory safety guarantees of traditional planning methods while incorporating the adaptability of data-driven approaches.}
    \item {To enable zero-shot scene understanding, we leverage one-shot exemplars and chain-of-thought prompting during the induction and deduction phase of MLLMs.}
    \item {We introduce a CVAE to model the relationship between navigation hyperparameters and natural language descriptions. This allows the system to learn expert-level parameter tuning strategies with textual guidance and generalize them to new scenarios.} 
    \item {Extensive experiments demonstrate that our framework outperforms existing navigation baselines in terms of safety, efficiency, comfort, robustness, decision transparency and human satisfaction level.}
\end{itemize}

\section{Related Work}
\subsection{Large language Model for Navigation}
Recent advances in MLLMs~\cite{GPT4o, QWEN, Kimi, Llama, qiu2025psn} have demonstrated remarkable performance and generalization capabilities in vision-language tasks. Their abilities in scene understanding, natural language interaction, and behavior interpretation have been increasingly applied in the field of robotics~\cite{shah2023lm, wen2024road, llmtase}. For navigation, previous works mostly integrate LLMs/MLLMs into path planning module. SayCan~\cite{brohan2023can} does high-level path planning with LLM. GPT-DRIVER~\cite{mao2023gpt} replaces traditional costmap with LLM. Wang \etal ~\cite{wang2024llm} employ LLM to plan and invoke the base motion module. Several works~\cite{yu2023l3mvn, shi2024lfenav} utilize LLM to find the relevant frontier in exploration task. LLaDA~\cite{li2024driving} obtains zero-shot traffic rulers information from LLM. NavCoT~\cite{lin2025navcot} trains the LLM to generate reasoning for navigation decisions. LLM-Copilot~\cite{qiao2024llm} enriches brief instructions given by humans through LLM. VLM-Social-Nav~\cite{song2024vlm} modifies socially compliant robot behavior with MLLM. 

\begin{figure}[t]
    \centering
    \includegraphics[width=1 \linewidth]{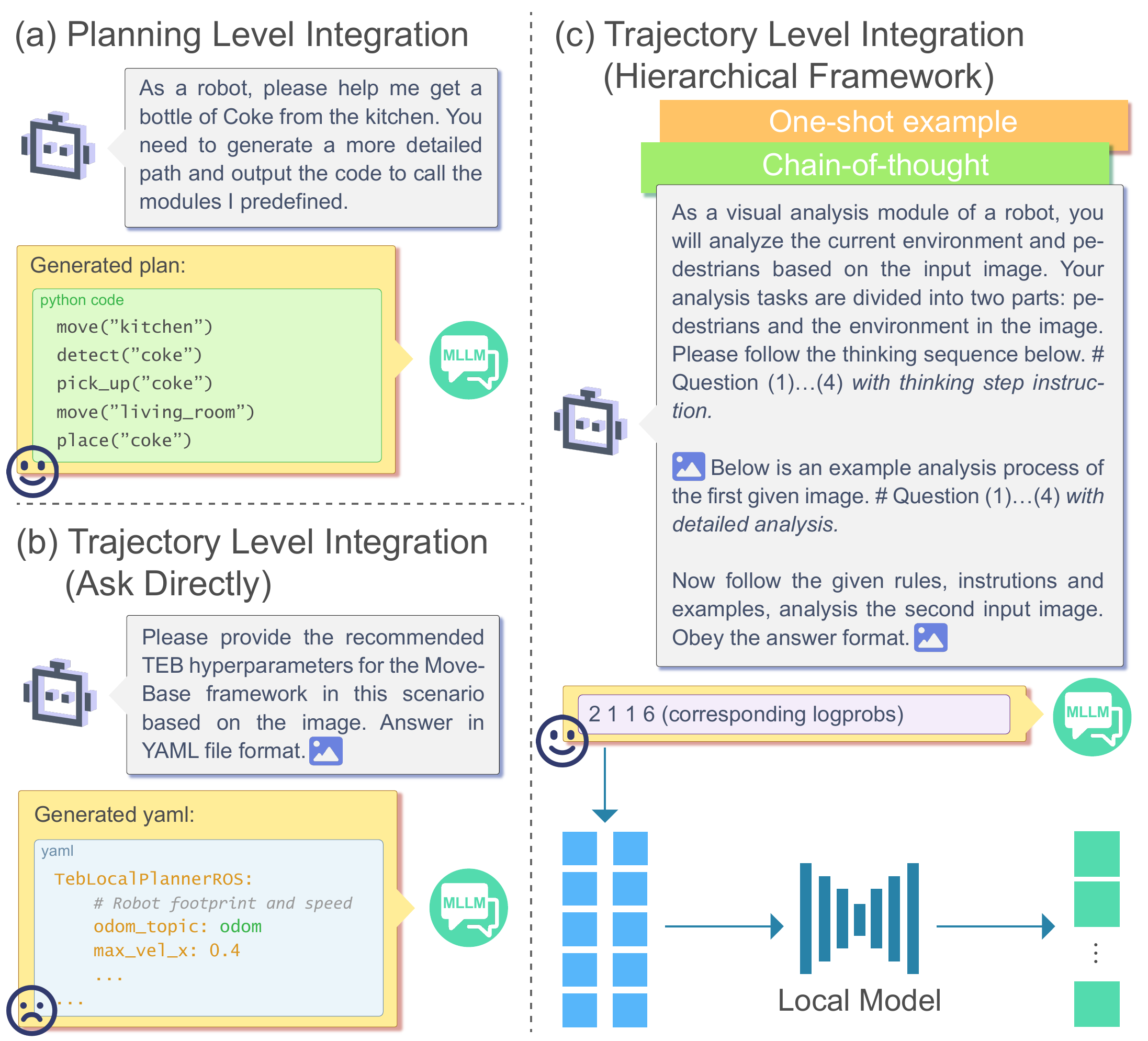}
    \caption{Comparison for LLM integration part in navigation. (a) The zero-shot and reasoning capabilities of LLMs have led to their success at the planning level. (b) Due to task misalignment, directly applying LLMs at the trajectory level often yields unsatisfied results. LLMs tend to output fixed hyperparameter values learned from the training corpus, rather than adapting to the current scene. (c) In our approach, the MLLM serves as an upstream module at the trajectory level by providing scene understanding, while a lightweight downstream model is trained to infer adaptive hyperparameter adjustments. }
    \label{fig2:llm}
\end{figure}

The success of previous works have demonstrated the zero-shot capabilities of LLMs/MLLMs and the potential of their textual explanations to support interpretable behavior in robotic navigation. Therefore, in our framework, we also leverage MLLMs to enhance the generalizability and decision transparency of navigation. However, fully end-to-end architectures such as vision-language-action(VLA) model or vision-language-navigation(VLN) model still face several limitations. These include the lack of collision-free guarantees, reduced decision interpretability, and degraded generalization due to underutilized language components. Aside from VLA architectures, most existing methods incorporate LLMs only at the planning level, as illustrated in Fig.~\ref{fig2:llm}. To enable more continuous and adaptive behavior modulation, we explore the integration of MLLMs into the local planner module. However, relying directly on MLLMs to determine navigation hyperparameters based on scene inputs is often inefficient. This is because, although many forms of intervention can influence the preference of LLMs/MLLMs toward specific features, it remains challenging to overcome their strong prior biases~\cite{chendanqi}.

Considering the weak coupling between perception and control, we adopt a hierarchical local planning architecture. Within this framework, an MLLM is employed to provide zero-shot scene descriptions, which are then used as conditional inputs for a lightweight model that dynamically adjusts navigation hyperparameters in real time.

\subsection{Data-Driven Hyperparameter Modulator for Optimization-Based Planners}
Conventional planning methods~\cite{DWA,TEB,gerkey2008planning,liuzhetase} have demonstrated strong performance in industrial robotic applications. However, for service robots operating in dynamic environments characterized by frequent context switching and human interference, such fixed strategies often prove insufficient. Some works~\cite{brock1999high, hong2015modified, missura2019predictive, teso2019predictive, dobrevski2020adaptive, chang2021reinforcement, pan2023d} focus on making up for the shortcomings of traditional DWA~\cite{DWA}, such as long-horizon planning~\cite{brock1999high} and overcoming dynamic interference~\cite{missura2019predictive}. Patel \etal \cite{patel2021dwa} reformulate the DWA factor computation and employ a neural network to directly predict the optimal control command, replacing the traditional cost-based selection. Chang \etal \cite{chang2021reinforcement} adjust the DWA factor configuration and adopt a tabular Q-learning approach to select suitable weight combinations for the cost function.

Although these approaches address various limitations of optimization-based planners while preserving the theoretical guarantees for collision avoidance, they still suffer from limited adaptability and may introduce new issues such as oscillations and robustness problems. The SOTA work DADWA~\cite{dadwa} introduces a neural network to predict the weights of the DWA cost function, enabling safe local navigation. To handle dynamic obstacles, the method incorporates a short sequence of observations, allowing the network to capture obstacle motion patterns and adapt the DWA weights accordingly. The network is trained by reinforcement learning (RL) in the simulation environment.

Inspired by the integration of data-driven and optimization-based planning approaches, we aim to develop a data-driven adaptive hyperparameter modulator. In practical applications, we observe that the SOTA method DADWA~\cite{dadwa} encounters several limitations. The presence of a significant sim-to-real gap often results in degraded performance when deployed in real-world scenarios. Moreover, the reward function used in RL may not align well with the principles of human–robot interaction, especially in the context of pedestrians or passengers interacting with the robot~\cite{gao2022evaluation}. Additionally, DADWA does not distinguish between different types of obstacles in its input representation. In contrast, most modern service robots are usually equipped with cameras, enabling the extraction of rich scene semantics from visual inputs, which can be leveraged to improve navigation decisions.

Motivated by these limitations, our method adopts a generative modeling framework that learns human expert tuning strategies from real-world data to acquire socially acceptable hyperparameter modulation capabilities. Furthermore, we incorporate a MLLM to generate real-time scene descriptions from visual input streams.

\section{Method}
\subsection{Problem Formulation}
Based on the analysis of prior works in the previous section, we formulate the hyperparameter modulator as a conditional generation problem:

\begin{equation}
    \mathcal{H}^{t} = f_{\theta}(s^{t:t-t_w},\epsilon^{t}_{(n)}),\label{eq:prob}
\end{equation}
where $\mathcal{H}^{t} \in \mathbb{R}^{N}$ are $N$ selected hyperparameters for each planner at time $t$. $f_{\theta}(\cdot,\cdot)$ is the conditional generator. The scene-description condition $s^{t:t-t_w}$ is acquired from MLLM. $ \epsilon^{t}_{(n)}$ is the noise variable for multi-model generation method and $n$ is the number of samples. To be more specific, we train LE-Nav as a conditional variational autoencoder, which will be discussed later.

\begin{figure*}[t]
  \centering
\includegraphics[width=1.0\linewidth]{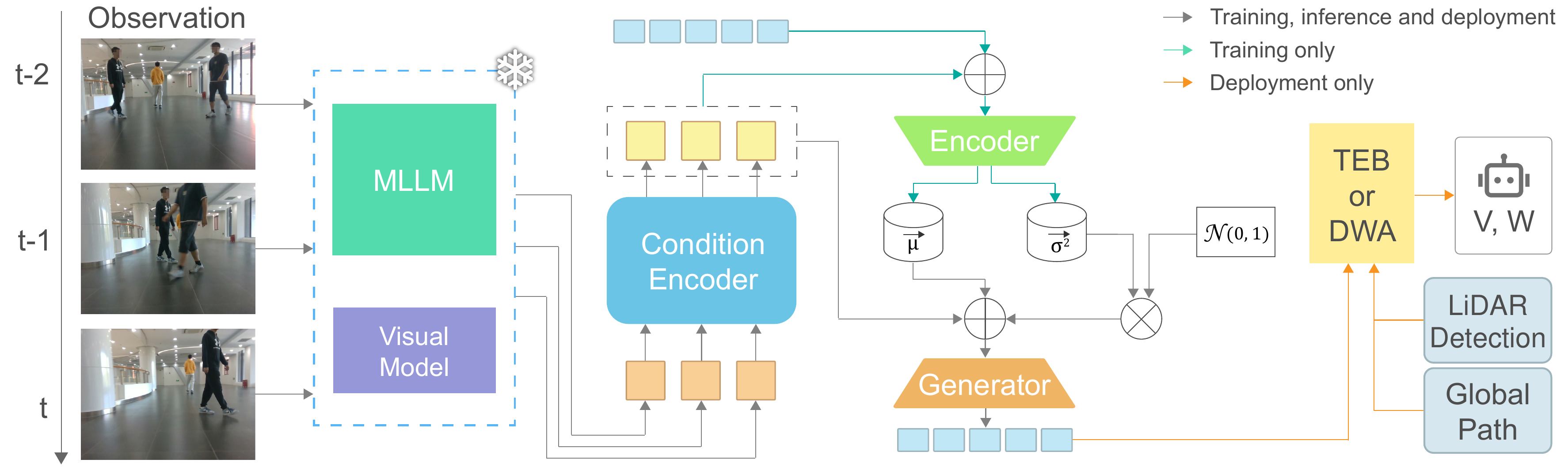}
        \caption{Overall framework. Scene context is first obtained from the current observation using the MLLM and visual model. It is then encoded by a transformer-based condition encoder. The hyperparameter encoder is used only during the training phase. The generator is responsible for decoding from the Gaussian latent space based on the encoded conditions.}
        \label{fig3:overall}
\end{figure*}

The overall framework is illustrated in Fig.\ref{fig3:overall}. We begin by briefly reviewing the timed elastic band (TEB) and dynamic window approach (DWA) planners in Sec.\ref{method:rev}. Next, we propose two prompting strategies to obtain stable and accurate scene analysis from the MLLM (Sec.\ref{method:MLLM}). Finally, we train the hyperparameter modulator using a CVAE, as described in Sec.\ref{method:cvae}.

\subsection{Rethinking Hyperparameter Modulation}
\label{method:rev}
\subsubsection{Brief Review of TEB}
TEB~\cite{TEB} formulates local trajectory planning as a constrained nonlinear optimization problem over a sequence of poses $\mathbf{x}_i = (x_i, y_i, \theta_i)$ and time intervals $\Delta T_i$. The goal is to minimize a weighted sum of costs:
\begin{align}
\mathcal{C}_{\text{total}} &= \sum_{i=1}^{N} \Bigl(
w_{\text{goal}} \cdot \mathcal{C}_{\text{goal}}(\mathbf{x}_N) +
w_{\text{smooth}} \cdot \mathcal{C}_{\text{smooth}}(\mathbf{x}_i, \mathbf{x}_{i+1}) \notag \\
&\quad + w_{\text{obs}} \cdot \mathcal{C}_{\text{obstacle}}(\mathbf{x}_i) +
w_{\text{dyn}} \cdot \mathcal{C}_{\text{dynamic}}(\mathbf{x}_i, \Delta T_i) \notag \\
&\quad + w_{\text{time}} \cdot \mathcal{C}_{\text{time}}(\Delta T_i)
\Bigr), \label{eq:teb}
\end{align}
where $\mathcal{C}_{\text{goal}}$ penalizes the final pose deviation from the global goal, and
$\mathcal{C}_{\text{smooth}}$ encourages trajectory smoothness, and $\mathcal{C}_{\text{obstacle}}$ penalizes proximity to obstacles, and $\mathcal{C}_{\text{dynamic}}$ ensures compliance with the robot’s kinematic and dynamic constraints, and $\mathcal{C}_{\text{time}}$ ensures temporal feasibility and speed regulation. The corresponding $w$ are the potential targets of LE-Nav for TEB.

\subsubsection{Brief Review of DWA}
DWA~\cite{DWA} selects the optimal control input $(v, \omega)$ by evaluating a set of feasible velocity commands within a dynamically constrained search space. The goal is to maximize the objective function:
\begin{align}
    G(v, \omega) &= w_{\text{head}} \cdot \mathcal{G}_{\text{head}}(v, \omega) + w_{\text{clear}}  \cdot \mathcal{G}_{\text{clear}}(v, \omega) \notag \\
    &\quad+ w_{\text{vel}} \cdot \mathcal{G}_{\text{vel}}(v), \label{eq:dwa}
\end{align}
where $\mathcal{G}_{\text{head}}$ measures the alignment with the goal direction, and $\mathcal{G}_{\text{clear}}$ quantifies the minimum distance to surrounding obstacles, and $\mathcal{G}_{\text{vel}}$ encourages higher translational speed. The corresponding $w$ are the potential targets of LE-Nav for DWA.

\subsubsection{Objective Hyperparameter}
In ROS~\cite{ROS}, both the TEB and DWA planners are well-engineered within the MoveBase package, where the global and local planning components are systematically organized. Although the implementations of TEB and DWA differ slightly from those described in Eq.~\eqref{eq:teb} and Eq.~\eqref{eq:dwa}, they offer enhanced robustness. Building upon this framework, we employ dynamic reconfiguration to enable real-time online hyperparameter tuning.

Based on Eq.~\eqref{eq:teb} and Eq.~\eqref{eq:dwa} and expert knowledge from human practitioners, we select the following eight key hyperparameters for TEB: \textit{max\_vel\_x}, \textit{max\_vel\_theta}, \textit{acc\_lim\_x}, \textit{acc\_lim\_theta}, \textit{weight\_max\_vel\_x}, \textit{weight\_acc\_lim\_x}, \textit{weight\_acc\_lim\_theta}, \textit{weight\_optimaltime} and eight key hyperparameters for DWA: \textit{max\_vel\_x}, \textit{max\_vel\_theta}, \textit{acc\_lim\_x}, \textit{acc\_lim\_theta}, \textit{path\_distance\_bias}, \textit{goal\_distance\_bias}, \textit{occdist\_scale}, \textit{forward\_point\_distance}. In the case of DWA, when updating hyperparameter \textit{max\_vel\_x}, we additionally synchronize the value of \textit{max\_vel\_trans} to be consistent with \textit{max\_vel\_x}. During the deployment, the \textit{inflation\_radius} of global costmap is also recorded and learned as it is closely related to the local planner. Therefore, both TEB and DWA has nine hyperparameters $\mathcal{H} \in \mathbb{R}^{9}$ for training and learning.

\subsection{Zero-Shot Scene Understanding via MLLM}
\label{method:MLLM}
As analyzed in Section 2.1 and illustrated in Fig. 2, during trajectory level integration, directly prompting a MLLM without end-to-end fine-tuning tends to yield hyperparameter suggestions closely aligned with the defaults recommended by the original toolkit. This behavior stems from experience-driven generalization and limits the model’s ability to adapt to specific scenarios. On the other hand, an end-to-end training strategy often leads to action outputs that lack formal guarantees of collision-free performance. To address these limitations, we adopt a hierarchical framework: the MLLM is tasked with generic scene understanding and provides a structured condition tensor $s^{t}$ to guide downstream hyperparameter generation. The MLLM is thus held accountable for the reasoning that supports both its interpretation of the scene and the resulting behavior of the navigation system, thereby improving decision transparency.

In this section, we detail the design of the induction and deduction phases of our MLLM-based reasoning. The complete prompt templates are available in our open-source repository. When constructing prompts, we address the following three questions: (1) How should the MLLM's output be formatted as input to the downstream CVAE model? (2) How can we enforce a standardized output structure and rules? (3) How can we improve the accuracy of the MLLM's responses?

\subsubsection{Output Intention of MLLM}
For Question (1), a direct textual scene description often introduces ambiguities in structure, vocabulary, and length, leading to increased model size and data requirements for training the downstream generation model. We observe that human experts typically evaluate an environment along several key dimensions when tuning hyperparameters. Inspired by this, we structure the MLLM’s output as a numerical matrix that captures these dimensions. Specifically, we decompose the scene into two components. Foreground (dynamic obstacles, such as pedestrians): The MLLM is prompted to assess whether pedestrians are present, their density, movement direction, and proximity to the robot. Background (environmental complexity): From a driver’s perspective, the MLLM must evaluate how challenging the environment is for safe navigation, including factors such as spatial constraints and potential collision risks.

After answering these five dimensions, the MLLM produces a structured ``scene rating" (as shown in Fig.~\ref{fig2:llm}), which we use as the conditional tensor input to the downstream policy model. Moreover, many MLLMs also provide a token-level log probability score, which serves as a confidence estimation. This score can optionally be used to enrich the downstream model’s input with uncertainty awareness.

\subsubsection{Induction Phase}
For Question (2), we adopt an induction strategy based on one-shot exemplars. Prior studies~\cite{kaplan2020scaling} have shown that providing an LLM with a few representative input-output pairs significantly improves output quality by demonstrating the desired reasoning pattern.

In the induction phase, we design a clear and explicit rule set, accompanied by a detailed example illustrating how a human expert would reason through the five scene-assessment questions discussed earlier. This exemplar allows the MLLM to generalize the reasoning process to arbitrary scenarios under the same rule framework, leveraging both the provided demonstration and its pretrained knowledge.

To further ensure consistency and usability, we standardize both the input and output formats using structured templates. Specifically, we explicitly define the output schema, enabling the MLLM to generate outputs that can be directly parsed and consumed by downstream models without complex post-processing.

\subsubsection{Deduction Phase}
For Question (3), we employ a chain-of-thought (CoT) prompting strategy. Empirical evidence~\cite{wei2022chain} in engineering practice suggests that encouraging the model to “think step-by-step” by explicitly decomposing complex tasks into intermediate reasoning steps significantly improves its performance on tasks requiring logical deduction.

Accordingly, in the deduction phase, we decompose the task into sub-questions and guide the MLLM to perform step-by-step reasoning. This not only improves the accuracy and robustness of the final output but also makes the reasoning process interpretable and auditable. Such interpretability is crucial in safety-critical applications like navigation, where transparent decision-making helps build trust and enables downstream systems to validate or override decisions when necessary.

Despite the rapid progress of MLLMs, their performance on certain fine-grained tasks such as check numbers remains suboptimal. For instance, even when provided with a rule-based heuristic such as “near objects appear larger than distant ones,” along with exemplars and chain-of-thought prompting, current models still struggle to estimate accurately human proximity in complex scenes. To address this limitation, we incorporate an auxiliary visual model
~\cite{yolov11} to assist in this specific subtask. We anticipate that with future advancements in MLLM capabilities, such visual assistance may eventually become unnecessary, enabling MLLMs to perform complete scene assessments independently.

\subsection{Hyperparameter Generation via CVAE}
\label{method:cvae}
The hyperparameter modulator is designed to bridge the gap between MLLM outputs and hyperparameter configurations. As formulated in Eq.~\eqref{eq:prob}, we cast this task as a CVAE problem, where the generator learns to produce expert-level hyperparameters conditioned on text descriptions provided by the MLLM.

As illustrated in Fig.~\ref{fig3:overall}, the hyperparameter modulator consists of three components: a condition encoder, an encoder, and a generator. The condition encoder takes as input the current frame and a short temporal history of the scene description $s^{t:t-t_w}$, as extracted in the previous stage. Given the sequential nature of the input, we employ a Transformer-based architecture for the condition encoder. In contrast, the encoder and decoder are implemented as lightweight MLPs to ensure computational efficiency. The encoder is only used during training to project expert-annotated hyperparameters into a Gaussian latent space. The decoder then reconstructs or generates hyperparameters from this latent space, conditioned on the features produced by the condition encoder.

To facilitate stable training and flexible deployment, all hyperparameters are normalized to the $[0,1]$ range via min-max scaling during training and rescaled back to their original units during inference. This normalization brings two benefits. (1) Balanced loss contribution: Hyperparameters typically vary in scale. Normalization ensures that each dimension contributes equally to the reconstruction loss, improving learning stability. (2) User personalization: For user-adjustable hyperparameters (e.g., maximum speed) that do not require expert knowledge, the normalized output can be easily mapped to user-preferred ranges. The model learns the relative intent (e.g.,``faster" vs. ``slower") rather than absolute values, enabling user-aware adaptation.

The overall loss function consists of two terms. The Kullback–Leibler (KL) divergence, which regularizes the latent space by aligning the encoder’s output with a Gaussian prior. The mean squared error (MSE), which quantifies the reconstruction loss between the generated and ground-truth hyperparameters conditioned on the textual context. The complete loss formulation of this CVAE~\cite{CVAE} framework is given as:

\begin{align}
    L(s^{t:t-t_w},\mathcal{H}^{t};\theta,\phi ) 
    &= -D_{KL}\left(
        q_{\phi}(z^{t}|s^{t:t-t_w},\mathcal{H}^{t}) \right. \notag \\
    &\quad \left\| 
        \left. p_{\theta}(z^{t}|s^{t:t-t_w}) \right) \right. + \frac{\gamma}{N} \notag \\
    &\quad  \sum_{n=1}^{N} \left\| 
        p_{\theta}(\mathcal{H}^{t}|s^{t:t-t_w},z^{t}_{(n)}) - \mathcal{H}^{t} 
        \right\|_{2}^{2}, \label{loss}
\end{align}
where $\theta$ and $\phi$ are learned parameters for networks. $z^{t}_{(n)} = f_{\phi}(s^{t:t-t_w},\mathcal{H}^{t},\epsilon^{t}_{(n)})$ is latent variable for CVAE, where $\epsilon^{t}_{(n)} \sim \mathcal{N}(0, 1)$. $\gamma$ is the loss balance weight.

The scene description $s$ is critical for accurate hyperparameter generation. However, in real-world deployment, packet loss from the MLLM is often unavoidable due to network latency, server congestion, or other system-level issues. To enhance robustness, we introduce a random packet loss augmentation strategy during training. Specifically, we simulate packet loss by randomly dropping frame features and masking them during attention computation. The model then computes attention over the remaining valid frames to construct the conditional representation. This strategy improves the model’s resilience to packet loss through both data augmentation and selective attention masking, enabling more reliable hyperparameter generation under imperfect communication conditions.

Unlike prior frameworks in action generation such as DP~\cite{zhao2023learning} and ACT~\cite{chi2023diffusion}, we do not adopt chunking or smoothing strategies. This decision is motivated by two key factors. First, the nature of our conditional input $s^{t:t-t_w}$ differs fundamentally. In contrast to high-frequency visual inputs used in other research domains, our condition originates from MLLM outputs, which are inherently low-frequency and delayed. This latency necessitates more immediate system feedback. Moreover, packet loss in MLLM transmissions can introduce errors that accumulate during smoothing, degrading performance. 

Second, our framework generates hyperparameters, rather than directly generates discrete actions. As a result, smoothing is unnecessary. In our hierarchical architecture, the hyperparameter generator serves as a high-level controller, dynamically modulating the parameters of underlying local planners (e.g., DWA or TEB) in real time. Importantly, these hyperparameter adjustments do not induce abrupt changes in motion, as the effect is absorbed by the inherent continuity of the planner's dynamics. This design enables the system to flexibly adapt to dynamic environments and user preferences, while maintaining smooth and stable motion through hierarchical control and the intrinsic properties of the motion planner.

\section{Experiment}
\subsection{Implementation Details}
\subsubsection{Platforms}
\begin{figure}[t]
    \centering
    \includegraphics[width=0.7 \linewidth]{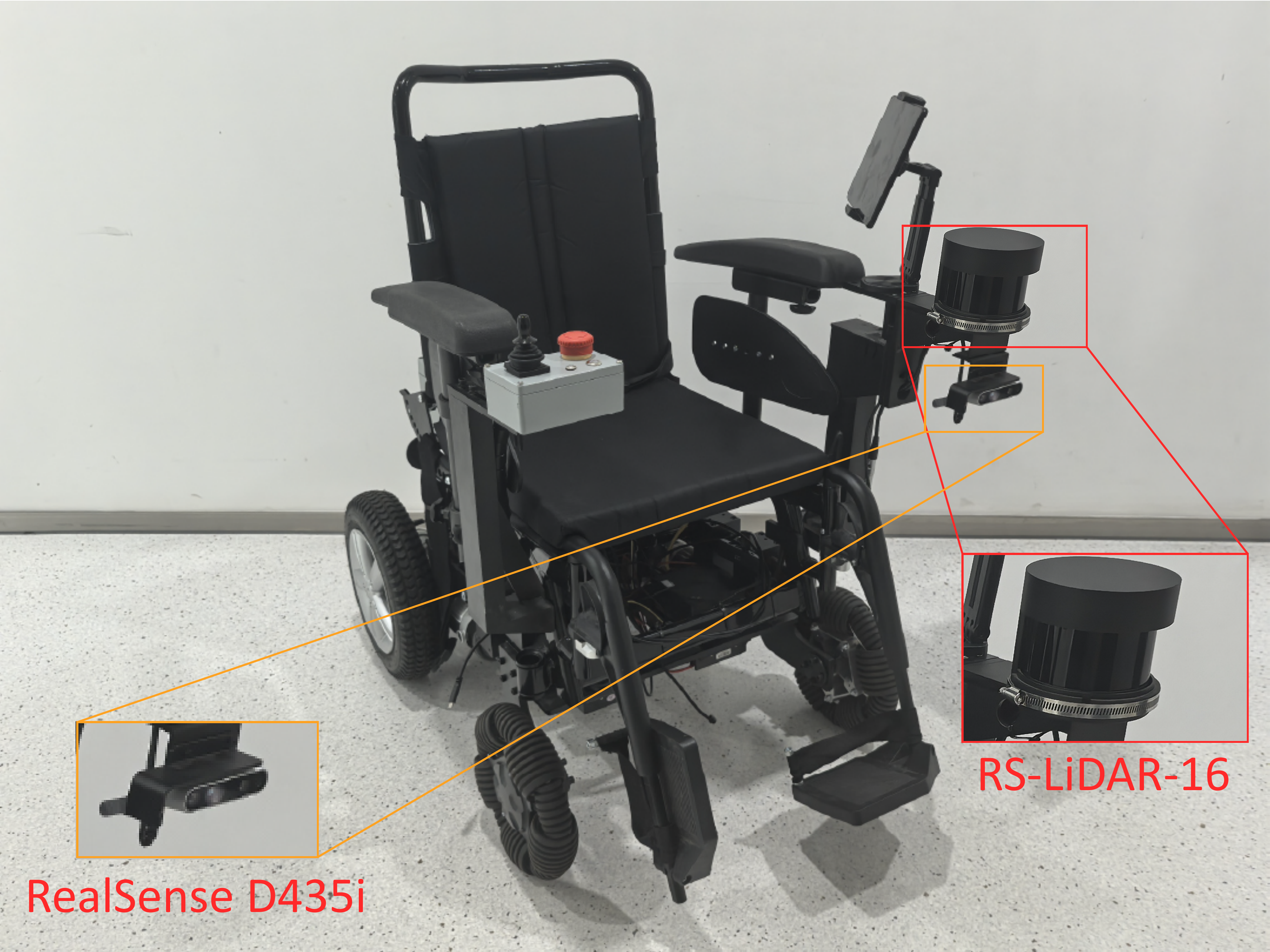}
    \caption{Experimental Platform. The image data stream captured by the onboard camera is utilized for MLLM analysis, enabling high-level semantic understanding of the environment. LiDAR data is primarily used for localization and fundamental obstacle detection.}
    \label{fig4:platform}
\end{figure}

In this work, we employ a typical service robot—a smart wheelchair—as the experimental platform for data collection, reproduction of existing method, navigation performance evaluation and interactive experience research. The smart wheelchair provides a unique testing scenario by involving both passenger interaction (onboard user commands and feedback) and pedestrian interaction (dynamic responses to people in the surrounding environment). As shown in Fig.~\ref{fig4:platform}, the custom-built wheelchair robot is equipped with key sensors including an RS-LiDAR-16 and an Intel RealSense D435i camera. To minimize the sim-to-real gap, we built a virtual replica of our smart wheelchair robot in the Gazebo simulation environment, enabling faithful reproduction of the previous approach DADWA \cite{dadwa}. 

\subsubsection{Data Collection and Deployment}
In Sec.~\ref{method:cvae}, we formulate hyperparameter tuning as a conditional generation problem. During the training phase, human experts manually adjusted the parameters of both TEB and DWA planners for different navigation scenarios. Throughout the process, we recorded a low-frequency stream of image data (0.5Hz) along with the corresponding selected hyperparameters using ROS bag files. The expert tuning process followed a combination of subjective judgment and objective criteria. The primary objective was to ensure the safety of both the robot and nearby pedestrians. Given that, the expert aimed to enable successful navigation in each specific scenario, followed by minimizing traversal time, aligning the path with the operator’s intuitive expectations, and finally, ensuring ride comfort.

For the TEB planner, we collected a total of 53 sequences, with 44 used for training and 9 for validation. For the DWA planner, 56 sequences were collected, of which 47 were used for training and 9 for validation. Each sequence lasts between dozens of seconds and a few minutes. This diverse set of sequences captures a wide range of navigation scenarios, providing a solid foundation for conditional generation modeling.

Our method is highly resource-efficient and requires minimal GPU memory, making training feasible with less than 0.5 GB of memory ($\text{batch size} = 16$). We trained the hyperparameter tuning models for both the TEB and DWA planners for 8000 epochs, using a learning rate of 0.001 with the AdamW optimizer. More implementation details of training hyperparameter setup can be found in the \href{https://github.com/Cavendish518/LE-Nav}{Github link}. During navigation, the wheelchair robot was controlled by a laptop equipped with an Intel i5-12400F CPU and an NVIDIA GeForce RTX 3060 GPU.

\subsection{Hyperparameter Generation Performance} 
\label{exp:MLLM}
\subsubsection{Generation Performance}
To evaluate the hyperparameter generation performance, we adopt a Top-\textit{k} strategy and introduce two metrics to assess whether LE-Nav has learned the human expert’s ability in hyperparameter tuning. The Top-\textit{k} strategy refers to generating k sets of hyperparameters per validation instance. Each of the k results is compared to the expert-provided hyperparameters using percentage error, and the average error is computed across all hyperparameters for each planner. Top-\textit{Err} denotes the lowest average percentage error among the k samples, while Mean-\textit{Err} indicates the mean of the average percentage errors across the k samples.

Although LE-Nav primarily focuses on the framework itself rather than the specific choice of MLLM, we still compare the results using three mainstream MLLMs as conditional inputs: GPT-4o, Qwen-VL-Max, and Kimi-latest. Among them, GPT-4o and Kimi-latest provide log-probabilities (logprobs) to support conditional variational inference, while Qwen-VL-Max from official platform requires padding to meet the CVAE input format. As MLLMs continue to advance, their outputs will become faster and more accurate, providing better contextual conditions for downstream generative models and ultimately leading to improved navigation performance under our framework.

As shown in Tab.~\ref{tab:1}, regardless of which MLLM is used as the conditional input, LE-Nav keeps the percentage error within 10\%, demonstrating a hyperparameter tuning capability comparable to that of human experts. Among the models, GPT-4o yields the most accurate outputs, resulting in the lowest hyperparameter error. However, considering real-time constraints (response frequency of 0.5 Hz), we adopt Qwen-VL-Max as the conditional input in the subsequent navigation experiments, as it meets the required latency under our network conditions.

We further analyze the generation performance for TEB and DWA. Due to their underlying algorithmic differences illustrated in Eq.~\eqref{eq:teb} and Eq.~\eqref{eq:dwa}, the two planners exhibit distinct characteristics. Most hyperparameters in TEB demonstrate a relatively linear progression between conservative and aggressive settings. In contrast, DWA's hyperparameters tend to be more complex and interdependent, with stronger coupling effects. Therefore, the hyperparameter error of TEB is smaller than that of DWA.

These results confirm that our method behaves as expected, enabling real-time hyperparameter adaptation and laying a solid foundation for downstream navigation tasks.

\begin{table}[htbp]
\caption{Learning performance with different MLLMs condition under random packet loss. Results are reported with Top-\textit{k = 10}. \label{tab:1}}
\centering
\begin{tabular}{l | l | c | c}
\hline
\textbf{Planner} & \textbf{Input MLLM} & \textbf{Top \textit{Err}} & \textbf{Mean \textit{Err}} \\
\hline
\multirow{3}{*}{TEB} 
    & GPT4o & 3.60\% & 3.77\%  \\
    & Kimi  & 8.05\% & 8.14\% \\
    & QWEN  & 6.70\% & 6.86\%\\
\hline
\multirow{3}{*}{DWA} 
    & GPT4o & 3.84\% & 4.03\% \\
    & Kimi  &  9.12\% & 9.42\% \\
    & QWEN  & 4.16\% & 4.83\% \\
\hline
\end{tabular}
\end{table}

\subsubsection{Ablation Study}
In Sec.~\ref{method:cvae}, we introduce a random packet loss enhancement strategy during training to address inevitable packet loss in real-world deployment. The results of the ablation study are shown in Table~\ref{tab:2}. Compared to models trained without this enhancement, performance degrades significantly under packet loss, resulting in unreliable hyperparameter predictions and potential safety risks during navigation.

In particular, during real-world navigation, we observe that issues such as network latency and server bandwidth limitations often lead to the loss of the latest one or two frames. For commonly observed packet loss patterns, models that have not undergone our robustness-oriented training strategy exhibit nearly twice the error compared to those that have. Such a significant degradation in performance can be catastrophic for downstream navigation tasks.

\begin{table}[htbp]
\caption{Comparison of results under different training strategies. w/ and w/o indicate training with and without the random packet loss enhancement strategy and corresponding mask calculation, respectively.
Random denotes the validation mode with randomly dropped packets.
Latest refers to the mode where the latest frame is dropped during validation, while Latest Two refers to dropping the latest two frames. Results are reported in Top-\textit{Err} / Mean-\textit{Err} format. \label{tab:2}}
\centering
\resizebox{\linewidth}{!}{
\begin{tabular}{l | l | l | l | l}
\hline
\textbf{Planner} & \textbf{Enhancement} & \textbf{Random} & \textbf{Latest} & \textbf{Latest Two} \\
\hline
\multirow{3}{*}{TEB} 
    & w/ & 6.70\% / 6.86\% & 7.69\% / 8.10\% & 9.43\% / 9.88\%\\
    & w/o & 9.16\% / 9.67\% & 12.48\% / 12.75\% & 17.89\% / 18.23\% \\
\hline
\multirow{3}{*}{DWA} 
    & w/ & 4.16\% / 4.83\% & 7.45\% / 7.59\% & 9.95\% / 10.01\% \\
    & w/o & 5.73\% / 6.30\% & 12.70\% / 12.81\% & 15.86\% / 15.89 \\
\hline
\end{tabular}
}
\end{table}

\subsubsection{Expert Policy vs. LE-Nav}
Moreover, we have observed that LE-Nav may perform better than human experts as hyperparameter modulators. For example, the hyperparameters of various scenes for human experts are discrete. This is because human cognition of scenes is more like a classification problem, while the generative model can be loosely regarded as a regression problem. This makes the hyperparameters generated by LE-Nav more continuous. We can further observe this feature in the subsequent experiments on navigation in Sec.~\ref{exp:navvis}.

\subsection{Navigation Performance}
\label{exp:nav}
\subsubsection{Experiment Scenarios}
\begin{figure*}[t]
  \centering
\includegraphics[width=0.9\linewidth]{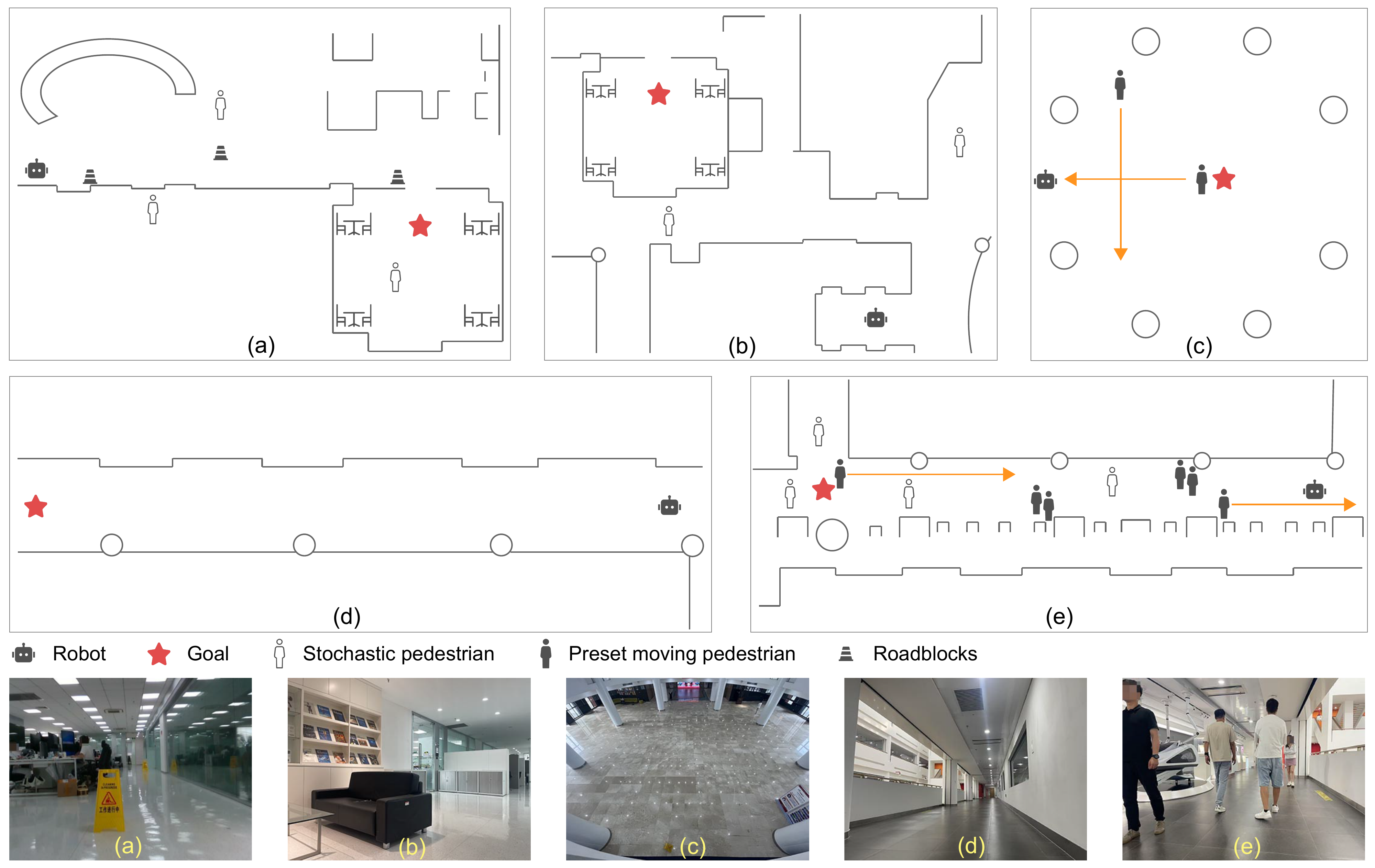}
        \caption{Real-world experiment maps and scenes. Stochastic pedestrians are non-design pedestrians that appear in very small quantities. Preset moving pedestrians are required to follow a fixed behavior. Roadblocks are temporary static disturbances. (a) Narrow Space with Unknown Obstacles. (b) Continuous Turns in Tight Corridors. (c) Dynamic Multi-Agent Crossing. (d) Static Path Efficiency. (e) High-Density Pedestrian Avoidance.}
        \label{fig5:map}
\end{figure*}

As shown in Fig.~\ref{fig5:map}, we selected five representative task scenarios to evaluate the navigation performance improvements achieved by LE-Nav. To ensure reproducibility and statistical rigor, the robot was required to repeatedly navigate from the same starting point to the same goal. Preset moving pedestrians followed identical motion patterns across trials to maintain consistency. The testing order of different methods was randomized, and each method was evaluated three times.

\textbf{Scenario (a) Narrow Space with Unknown Obstacles:} This scenario involves navigating through a confined environment containing static obstacles that are not present in the global costmap, posing a challenge for global-local planner coordination.

\textbf{Scenario (b) Continuous Turns in Tight Corridors:} The main difficulty in this scenario lies in a sequence of sharp turns combined with narrow spaces, requiring precise local planning.

\textbf{Scenario (c) Dynamic Multi-Agent Crossing:} A dynamic environment where multiple agents (e.g., pedestrians) move in intersecting paths, testing the system’s ability to handle social navigation.

\textbf{Scenario (d) Static Path Efficiency:} A relatively simple static environment designed to observe the efficiency and smoothness of the robot’s planned trajectories.

\textbf{Scenario (e) High-Density Pedestrian Avoidance:} A crowded scenario focused on evaluating the robot’s dynamic obstacle avoidance capability in the presence of dense pedestrian traffic.

In each scenario, we compared the following methods: TEB-Progressive, TEB-Conservative, DWA-Progressive, DWA-Conservative, DADWA~\cite{dadwa}, LE-Nav-TEB, and LE-Nav-DWA. Among them, the Progressive and Conservative variants of TEB and DWA refer to two distinct sets of hyperparameters carefully tuned by human experts across different scenarios, emphasizing efficiency and safety/comfort, respectively. To address the limitations of reinforcement learning, we trained DADWA in environments that structurally resembled scenarios (b), (c), (d), and (e), and introduced 10 simulated pedestrians based on the social force model to simulate dynamic interactions. To highlight the zero-shot generalization capability of our method, we ensured that, apart from the simplest static scenario (d), the training set did not contain any scene setups identical to those used during evaluation.

\subsubsection{Evaluation Metrics}
We choose the following metrics to evaluate the navigation performance.

\textbf{Failed to plan:} This metric includes cases of planning timeout or collision and is used to evaluate the method's ability to successfully complete a navigation task.

\textbf{Ego progress:} Defined as the time taken to complete the assigned navigation task, this metric reflects the efficiency of the navigation method.

\textbf{Risk rate:} Risk rate quantifies potential collision risks and is computed based on the time-to-collision ($TTC$) (calculated at 10 Hz). The basic risk rate is defined as the proportion of timestamps during navigation where the $TTC$ within the local costmap is infinite, indicating no imminent collision. Additionally, we report risk rates under relaxed safety thresholds, specifically for $TTC > 5s$ and $TTC > 2s$, to provide a more comprehensive assessment of safety performance.

\textbf{Acceleration and Jerk:} These metrics are used to objectively evaluate the comfort of the navigation method, where lower values typically indicate smoother and more comfortable trajectories.

Most previous methods only evaluated the success rate and efficiency. The expansion of the aforementioned metrics inevitably introduces trade-offs between them. Therefore, we set up a new score based on the above metrics to evaluate the navigation performance as a whole. This metric can uniformly evaluate the success rate, efficiency, comfort and safety of navigation, which is grounded in principles from human driving psychology~\cite{tsvetkova2024new}.

\begin{align}
    Score &= R_{\text{suc}} \cdot \big(\alpha_{1} T_{\text{norm}} + \alpha_{2} \text{Acc}_{\text{norm}} + \alpha_{3} J_{\text{norm}} \notag \\
    &\quad + \sum_{i=1}^{3} \beta_{i} R_{\text{safe}}^{(i)} \big), \label{score}
\end{align}
where $R_{\text{suc}}$ is the success rate and $R_{\text{safe}} = 1 - \text{Risk Rate}$. $T_{\text{norm}}$, $\text{Acc}_{\text{norm}}$ and $J_{\text{norm}}$ are obtained via min-max normalization across all compared methods. We set $\alpha_{1,2,3} = \{ 0.8, 0.1,0.1\} $ and $\beta_{1,2,3} = \{ 0.5,1.0,1.5\} $. We will discuss more about this newly proposed metric in supplementary materials.

\begin{table*}[h]
\caption{Navigation performance on five specific tasks. Failed to plan includes timeout and collision. The results of Ego progress, Risk Rate, Acceleration, and Jerk are obtained by calculating the average values of successful runs. Risk rates are reported as follows: risk rate within the local costmap / risk rate relaxed to $TTC > 5s$ / risk rate relaxed to $TTC > 2s$. \label{tab:3}}
\centering
\begin{tabular}{l | l | c | c | c | c | c | c}
\hline
\textbf{Experiment scene} & \textbf{Method} & \textbf{Failed to plan} & \textbf{Ego progress} & \textbf{Risk rate} & \textbf{Acceleration} & \textbf{Jerk} & \textbf{Score} \\
\hline
\multirow{7}{*}{Scenario (a)} 
    & TEB-progressive & 3 & - & - & - & - & 0.0 \\
    & TEB-conservative & 0 & 44.22 & 33.35\% / 28.32\% / 10.63\% & 0.14 & 8.86 & 0.7722 \\
    & DWA-progressive & 0 & 35.37 &  53.56\% / 31.65\% / 19.19\% & 0.38 & 9.74 & 0.7338 \\
    & DWA-conservative & 0 & 60.53 & 25.76\% / 18.86\% / 9.63\% & 0.28 & 9.73 & 0.6547\\
    & DADWA & 3 & - & - & - & - & 0.0 \\
    & \textbf{LE-Nav-TEB} & 0 & 34.35 & 40.45\% / 24.93\% / 12.53\% & 0.36 & 10.09 & \textcolor{blue}{0.7957} \\
    & \textbf{LE-Nav-DWA} & 0 & 34.59 & 42.93\% / 26.84\% /15.37\% & 0.34 & 10.29 & \textcolor{blue}{0.7739} \\
\hline
\multirow{7}{*}{Scenario (b)} 
    & TEB-progressive & 3 & - & - & - & - & 0.0 \\
    & TEB-conservative & 0 & 67.56 & 49.50\% / 38.39\% / 12.26\% & 0.29 & 9.25 & 0.6960 \\
    & DWA-progressive & 0 & 44.50 &  49.96\% / 37.56\% / 19.75\% & 0.37 & 10.43 & \textcolor{blue}{0.7313} \\
    & DWA-conservative & 0 & 91.53 & 62.54\% / 52.19\% / 18.38\% & 0.28 & 9.73 & 0.5145 \\
    & DADWA & 2 & 47.43 & 48.31\% / 38.39\% / 20.22\% & 0.39 & 10.76 & 0.2358 \\
    & \textbf{LE-Nav-TEB} & 0 & 55.08 & 52.34\% / 37.15\% / 15.56\% & 0.40 & 10.21 & 0.6975\\
    & \textbf{LE-Nav-DWA} & 0 & 48.11 & 44.81\% / 32.83\% / 17.37\% & 0.34 & 9.99 & \textcolor{blue}{0.7567}\\
\hline
\multirow{7}{*}{Scenario (c)} 
    & TEB-progressive & 0 & 11.93 & 9.94\% / 5.85\% / 4.76\% & 0.53 & 10.86 & 0.9020 \\
    & TEB-conservative & 0 & 19.35 & 9.47\% / 6.98\% / 2.93\% & 0.31 & 8.63 & 0.8432 \\
    & DWA-progressive & 0 & 11.19 & 10.29\% / 6.04\% / 4.39\% & 0.47 & 11.81 & \textcolor{blue}{0.9121}  \\
    & DWA-conservative & 0 & 25.22 & 6.73\% / 5.56\% / 3.56\% & 0.30 & 10.15 & 0.7524 \\
    & DADWA & 0 & 13.88 & 10.07\% / 6.86\% / 4.87\% & 0.38 & 9.82 & 0.8956 \\
    & \textbf{LE-Nav-TEB} & 0 & 14.86 & 8.44\% / 4.83\% / 3.34\% & 0.38 & 9.50 & 0.8970 \\
    & \textbf{LE-Nav-DWA} & 0 & 13.90 & 8.14\% / 7.58\% / 4.17\% & 0.31 & 8.96 & \textcolor{blue}{0.9129} \\
\hline
\multirow{7}{*}{Scenario (d)} 
    & TEB-progressive & 0 & 24.02 & 18.82\% / 6.22\% / 3.87\% &  0.33 & 9.90 & 0.9081 \\
    & TEB-conservative & 0 & 45.83 & 4.25\% / 2.98\% / 0.21\% & 0.23 & 8.31 & 0.8396 \\
    & DWA-progressive & 0 & 26.19 & 15.73\% / 1.39\% / 0.76\% & 0.35 & 10.39 & \textcolor{blue}{0.9115} \\
    & DWA-conservative & 0 & 53.72 & 5.82\% / 4.45\% / 0.87\% & 0.26 & 9.11 & 0.7633 \\
    & DADWA & 0 & 29.81 & 5.52\% / 3.31\% / 1.42\% & 0.34 & 10.58 & 0.8926 \\
    & \textbf{LE-Nav-TEB} & 0 & 26.79 & 17.81\% / 3.94\% / 1.54\% & 0.35 & 10.58 & 0.8935 \\
    & \textbf{LE-Nav-DWA} & 0 & 28.01 & 15.33\% / 0.24\% / 0.12\% & 0.34 & 10.29 & \textcolor{blue}{0.9082} \\
\hline
\multirow{7}{*}{Scenario (e)} 
    & TEB-progressive & 1 & 30.92 & 20.53\% / 12.42\% / 8.20\% & 0.41 & 12.03 & 0.5750 \\
    & TEB-conservative & 0 & 53.22 & 18.47\% / 15.21\% / 7.63\% & 0.24 & 8.57 & \textcolor{blue}{0.8003} \\
    & DWA-progressive & 2 & 30.92 & 22.83\% / 15.11\% / 11.58\% & 0.36 & 11.01 & 0.2850 \\
    & DWA-conservative & 0 & 71.49 & 17.63\% / 13.81\% / 5.83\% & 0.28 & 9.84 & 0.7065\\
    & DADWA & 1 & 34.61 & 23.41\% / 14.46\% / 9.40\% & 0.37 & 11.61 & 0.5600 \\
    & \textbf{LE-Nav-TEB} & 0 & 45.65 & 21.67\% / 13.93\% / 7.13\% & 0.30 & 9.42 & \textcolor{blue}{0.8238} \\
    & \textbf{LE-Nav-DWA} & 1 & 50.55 & 19.27\% / 14.23\% / 5.83\% & 0.31 & 9.91 & 0.5345 \\
\hline
\end{tabular}
\end{table*}

\subsubsection{Performance Analysis}
We analyze the advantages and disadvantages of different methods based on the metrics in Tab.~\ref{tab:3} and the characteristics of each scenario, followed by a summary of the overall performance.

For Scenario (a), in addition to the challenge of navigating through narrow spaces, unknown obstacles not represented on the global costmap pose significant difficulties for the local planner. Specifically, hyperparameters related to global trajectory adherence and other optimization objectives are subject to strict requirements. In such quasi-static, narrow-space scenarios, DWA-based methods exhibit higher tolerance to imperfect hyperparameter tuning compared to TEB, making them relatively more robust in this context. However, DADWA performs poorly in previously unseen scenarios, largely due to the lack of generalization capability in its framework. By leveraging the general scene understanding ability of large models, LE-Nav can reliably and accurately capture scene characteristics. This enables the hyperparameters to adapt dynamically based on real-time observations, allowing the system to operate safely and efficiently even in constrained environments. While this adaptation may introduce slight compromises in comfort, such trade-offs are negligible compared to the significant overall gains in comfort and efficiency. Notably, the adaptive mechanism allows our LE-Nav-TEB to achieve outstanding overall performance, highlighting the strong adaptability of our approach.

In Scenario (b), a series of sharp turns and narrow corridors pose considerable challenges for trajectory optimization in local planning. The confined space may exacerbate the sim-to-real gap, resulting in a relatively low success rate for DADWA. After hyperparameter adjustment using our method, both TEB and DWA achieve excellent success rates and safety levels within their respective algorithm families, while also striking a strong balance between efficiency and comfort.

Scenario (c) involves intersecting motions and serves as a representative benchmark for social navigation. Counterintuitively, conservative hyperparameter settings are not always preferable in dynamic environments. In scenarios with ample space for avoidance, a more progressive and assertive strategy may lead to better performance. Even though this type of complex, high-level semantic scenario is not included in the training set of LE-Nav, our method still demonstrates significant advantages.

For Scenario (d), which represents a relatively simple environment, all methods are capable of completing the task. Our approach further improves safety and comfort while maintaining high efficiency.

In Scenario (e), the dense crowd with limited clearance imposes significant challenges for local planning. In this case, TEB-based methods outperform DWA by a notable margin. Although DADWA still requires improvements in safety and comfort, it demonstrates competitive efficiency in the successful runs. Our LE-Nav-TEB clearly outperforms all other methods in terms of overall performance.

In summary, our approach exhibits two clear advantages across all scenarios:  \textbf{High adaptability:} Our method demonstrates strong adaptability, successfully completing navigation tasks across all testing environments through automatic hyperparameter adjustment. It provides significant advantages in safety, while also achieving a favorable trade-off between efficiency and comfort. In each method family, our variants consistently lead in performance. \textbf{Strong generalization capability:} Our approach generalizes well to unseen setups, as demonstrated by the impressive performance in Scenarios (a), (b), (c), and (e), all of which involve environments not included during training. These two advantages are also the motivation for us to establish LE-Nav.

\subsubsection{Navigation Visualization}
\label{exp:navvis}
\begin{figure}[htbp]
  \centering
\includegraphics[width=1.0\linewidth]{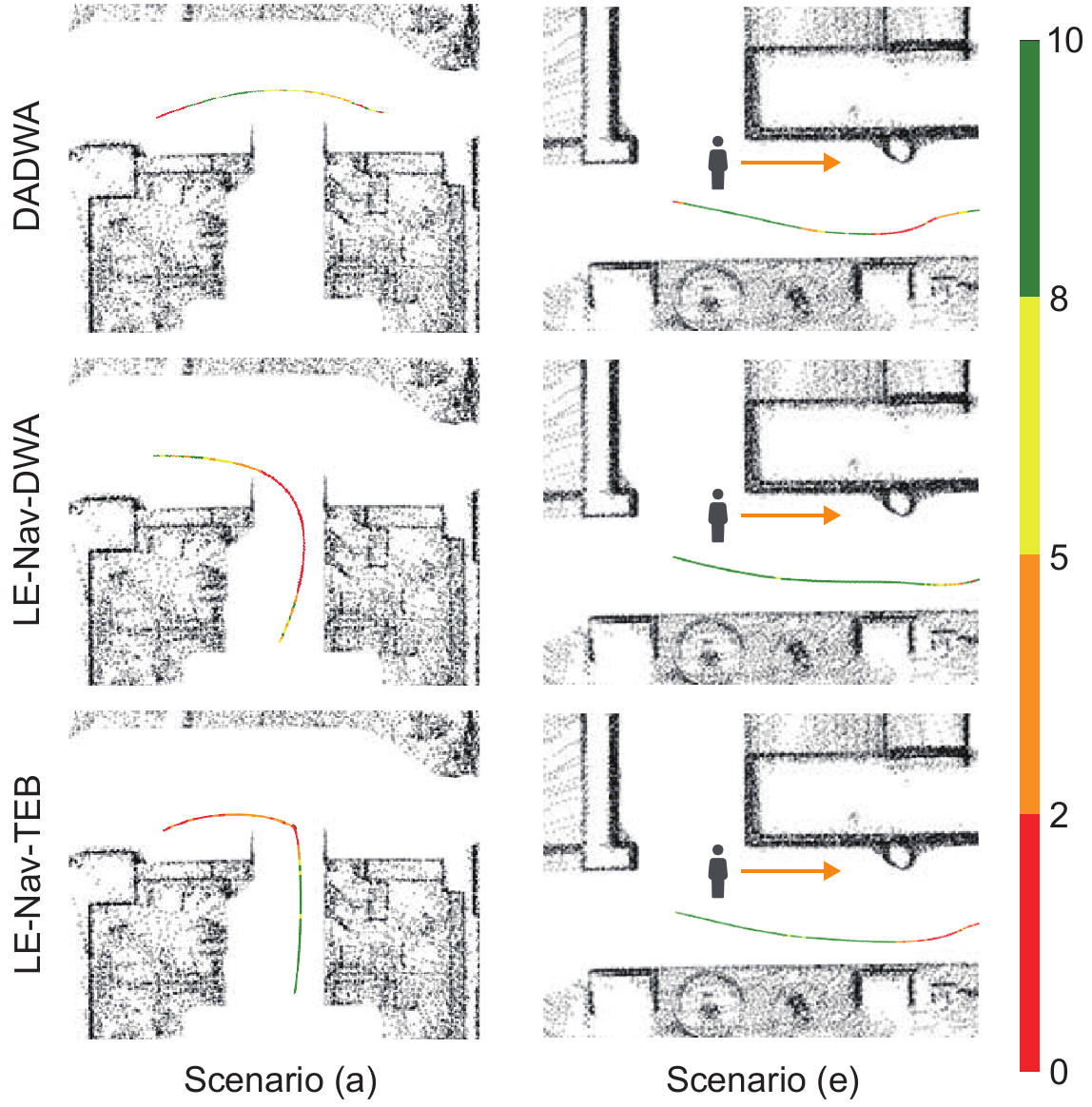}
        \caption{The $TTC$ along the trajectories is visualized using a color map, where $TTC$ values greater than 10 seconds and regions with no imminent collision risk are both mapped to the same color representing $TTC = 10$. Notably, TEB sometimes generates backward actions to adjust the robot's pose when passing through narrow gate.
}
        \label{fig:traj}
\end{figure}

As shown in the Fig.~\ref{fig:traj}, we further visualize and compare the local trajectories discussed in Sec. IV. For Scenario (a), we select the trajectory segment near the goal that involves passing through a narrow gate. We observe that DADWA often fails due to disturbances caused by the robot's weight in the real world and discrepancies in velocity tracking performance between simulation and reality. These factors cause DADWA to miss the optimal turning point in front of the narrow gate. In contrast, LE-Nav is trained on real-world expert demonstrations, where human experts implicitly account for such disturbances during parameter tuning. As a result, LE-Nav exhibits more robust performance.

For Scenario (e), we visualize the interaction trajectories near the goal where the robot encounters pedestrians. LE-Nav-TEB and LE-Nav-DWA both outperform DADWA in terms of safety (measured by $TTC$) and social acceptance (timing of avoidance). A closer analysis reveals that LE-Nav adapts its optimization priorities when facing close-range, oncoming pedestrians. Specifically, it tends to reduce the maximum velocity and acceleration limits to avoid sudden movements, and increases the weights of obstacle-related cost terms in the planner, shifting the focus away from efficiency and toward safety and comfort. By initiating avoidance earlier, LE-Nav produces trajectories that are not only safer but also more acceptable to both pedestrians and users. These results highlight the advantages of our proposed framework.

 We also use a \href{https://drive.google.com/file/d/1_XVsA-nbONcEre_OyEVM9BInMulYK7_r/view?usp=sharing}{\textcolor{blue}{video}} to further demonstrate the advantages of LE-Nav framework. This long-horizon navigation process lasted several minutes and had a significant change in hyperparameter strategy, which is also a common problem faced by service robots. In this video, we have visualized the most intuitive and common hyperparameter (the upper limit of speed) and have inversely transformed several representative outputs of MLLM into text and show the inevitable packet loss rate for MLLM. The reversed results from MLLM reveals the decision transparency of our methods. These two long-horizon demos further visualize the working logic of our method and verify the adaptive ability of our method. More analysis can be found in the supplementary materials.

% \begin{figure}[t]
%     \centering
%     \includegraphics[width=1.0 \linewidth]{vid.jpg}
%     \caption{A sample frame from the demo. The video illustrates how MLLM-based analysis indirectly influences hyperparameter adjustments, and how the generative model enables more continuous and real-time tuning compared to human experts.}
%     \label{fig:vid}
% \end{figure}

\subsection{User Study}
\label{exp:user}
\begin{table}[htbp]
\caption{Questionnaire targeting pedestrian and user perspectives. \label{tab:questionnaire}}
\centering
\resizebox{1.0\linewidth}{!}{
\begin{tabular}{l | l}
\hline
\textbf{Perspective} & \textbf{Questions (scored from 1 to 5)} \\
\hline
\multirow{2}{*}{Pedestrian} 
  & The wheelchair maintained a safe distance. \\
  & I felt comfortable sharing the space with the wheelchair in a public area. \\
\hline
\multirow{3}{*}{User}
  & I felt safe during the navigation process. \\
  & The wheelchair chose routes that I would find acceptable and appropriate. \\
  & The navigation felt smooth and did not cause discomfort. \\
\hline
\end{tabular}
}
\end{table}

\begin{figure}[htbp]
    \centering
    \includegraphics[width=1.0 \linewidth]{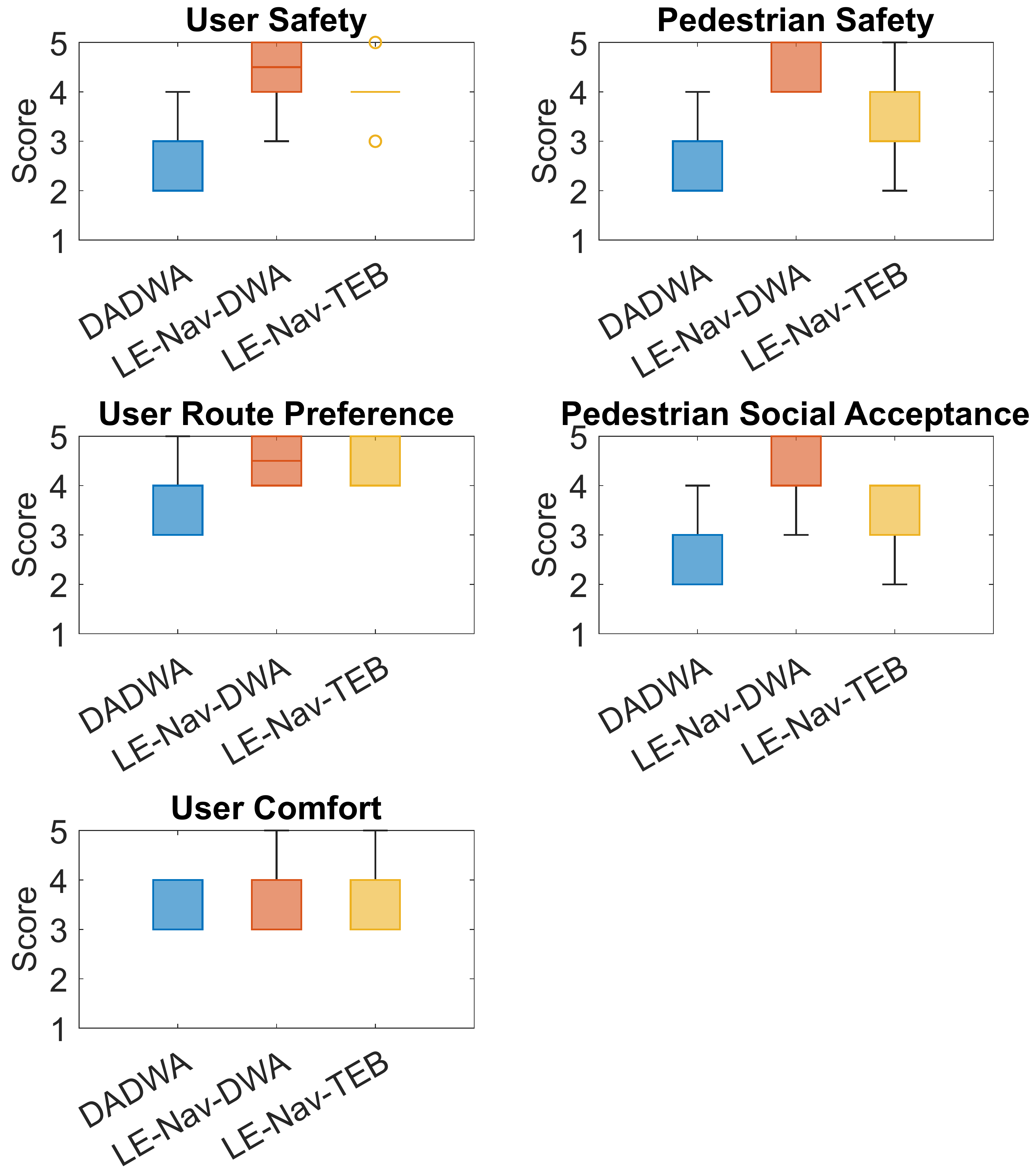}
    \caption{Scores from User study.}
    \label{fig6:userstudy}
\end{figure}

To validate the social acceptance of LE-Nav, we conducted a blind test involving ten groups of participants. The names of the methods have been hidden, and the three methods are randomly shuffled to ensure fair evaluation. In a corridor, users and pedestrians navigated and walked freely according to their own preferences without knowing each other’s destination, allowing for a fully natural interaction. Recall that in Sec.~\ref{method:cvae}, all hyperparameters were normalized. As a result, our framework supports personalized control by allowing users to adjust speed limits before the experiment begins, enabling customized ride experiences rather than relying solely on expert-designed models.

As shown in Tab.~\ref{tab:questionnaire}, we collected responses from two perspectives: psychological safety and social acceptance were assessed by pedestrians, while perceived safety, route expectation alignment, and user comfort were rated by users. The results are presented in Fig.~\ref{fig6:userstudy}.

The results show that our expert-data-driven generative framework outperforms the reinforcement learning-based approach across all five dimensions. For both pedestrians' and users' perceived safety, LE-Nav-DWA received the highest scores, followed by LE-Nav-TEB, and then DADWA, consistent with the safety metrics observed in the social navigation scenario (Scenario (c) in Tab.~\ref{tab:3}). For social acceptance, as rated subjectively by pedestrians, LE-Nav-DWA again scored highest, followed by LE-Nav-TEB and DADWA. For route alignment, LE-Nav-DWA and LE-Nav-TEB performed similarly and both outperformed DADWA. This discrepancy between pedestrian and user ratings on route alignment may be attributed to the fact that expert-tuned parameters were primarily designed from the user’s perspective, leading the LE-Nav model to learn more user-centered behaviors. In terms of user comfort, all three methods received positive evaluations, with LE-Nav showing a higher upper bound in performance.

Overall, the user study provides direct evidence of the social acceptance capability of the LE-Nav framework, and indirectly highlights the importance of decision transparency in enhancing the interaction experience between users, pedestrians, and the service robot.

\section{Conclusion}
In this work, we presented LE-Nav, a novel scene-aware and interpretable navigation framework that enables expert-level adaptive hyperparameter tuning for service robots operating in dynamic and unstructured environments. By decoupling scene understanding and hyperparameter generation, LE-Nav integrates MLLM reasoning with CVAE-based learning, allowing the system to generalize expert tuning strategies across diverse real-world scenarios. Through one-shot exemplars and chain-of-thought prompting, the MLLM generates semantically rich scene descriptions, which are then used by the CVAE to produce planner-specific hyperparameters. This two-stage architecture enhances decision transparency and leads to better navigation performance. Extensive experiments, including over a hundred real-world trials and a user study on a smart wheelchair platform, demonstrate that LE-Nav consistently outperforms state-of-the-art methods in terms of success rate, efficiency, safety, comfort, and perceived social acceptance. Overall, LE-Nav bridges the gap between data-driven adaptation and classical planning safety guarantees. Its ability to generate smooth and interpretable navigation behavior, combined with strong generalization to unseen scenes, makes it a practical and human-aligned solution for future service robotics applications.

Future work can be carried out from three aspects. First, in the direction of scene description. In addition to developing more effective prompts tailored to the suite of tasks in this work, we emphasize our contribution to building a full-stack navigation pipeline, where the MLLM functions as an upstream module whose capabilities can be progressively improved with advances in foundation models. In the future, MLLMs may acquire the capability to process real-time video streams, which will enable our framework to use faster and more continuous scene descriptions as conditional input. Second, improvements in planners. We will explore extending LE-Nav to a wider range of planners to further enhance overall navigation performance. In addition, we plan to improve the optimization objectives of existing planners by incorporating personalized user preferences, thereby fully leveraging the strengths of both MLLM and our local generative model. Finally, our framework could be extended to multi-robot systems, where shared scene understanding, cooperative decision-making, and multi-agent planning can be seamlessly integrated to support collaborative navigation tasks.

\section*{Acknowledgments}
We would like to thank the dozens of volunteers who participated in our navigation experiments. We also appreciate the valuable suggestions provided by Tianchen Deng, Ting Zhang and Chuan Cao for improving this work.

% {\appendix[Proof of the Zonklar Equations]
% Use $\backslash${\tt{appendix}} if you have a single appendix:
% Do not use $\backslash${\tt{section}} anymore after $\backslash${\tt{appendix}}, only $\backslash${\tt{section*}}.
% If you have multiple appendixes use $\backslash${\tt{appendices}} then use $\backslash${\tt{section}} to start each appendix.
% You must declare a $\backslash${\tt{section}} before using any $\backslash${\tt{subsection}} or using $\backslash${\tt{label}} ($\backslash${\tt{appendices}} by itself
%  starts a section numbered zero.)}

{\appendices
\section*{Appendix A: Performance Score}
\label{discuss score}
In Eq.~\eqref{score}, we propose a novel navigation performance metric that integrates four quantitative factors: success rate, efficiency, comfort, and safety. Moreover, experimental results in Tab. III demonstrate the scientific validity and practical effectiveness of our proposed index. Due to the differing optimization objectives of DWA (Eq.~\eqref{eq:dwa}) and TEB (Eq.~\eqref{eq:teb}) planners, each exhibits advantages in specific scenarios. In the five benchmark tasks we designed, the performance ranking of DWA-based and TEB-based methods aligns well with theoretical expectations: in scenarios where DWA is expected to perform better, DWA-series methods achieve significantly higher scores than TEB-based ones and vice versa.

\section*{Appendix B: Generalizability}
The four of five benchmark tasks in Sec.~\ref{exp:navvis} involve unseen environment setup during training of LE-Nav. Quantitative comparative experiments demonstrate the strong generalization capability of our method. To further validate this, we conducted extended long-term runs across diverse real-world scenarios. As shown in Fig.~\ref{fig7:gene}, we performed several hours of continuous navigation in café, lounge, platform with slope, and parking lot. Throughout these trials, our method consistently maintained collision-free and no-complaints performance, further confirming that LE-Nav achieves robust and reliable navigation across multiple environments.

\begin{figure}[t]
    \centering
    \includegraphics[width=1.0 \linewidth]{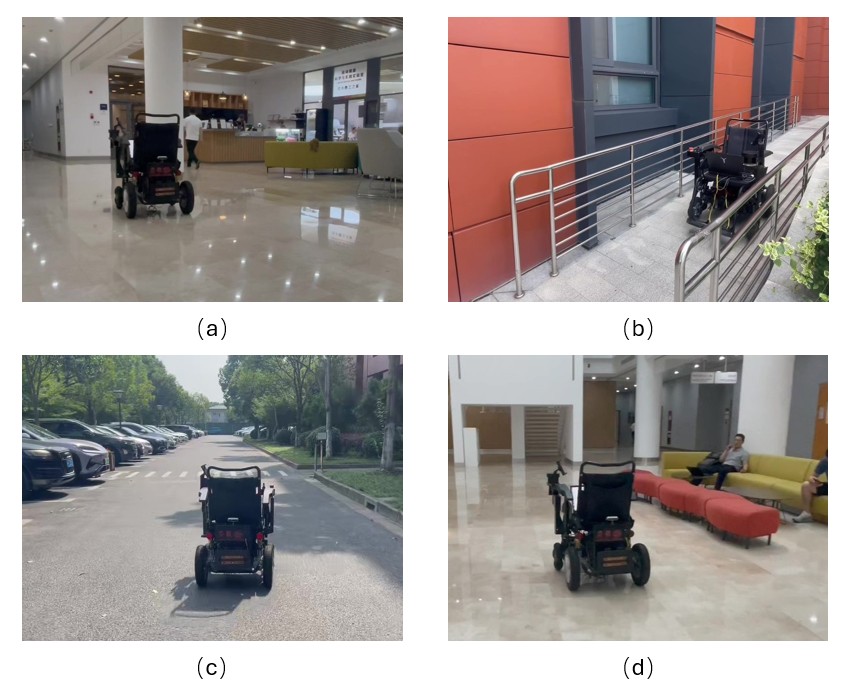}
    \caption{Generalization Test. (a) Café (b) Slope (c) Parking lot (d) Lounge.}
    \label{fig7:gene}
\end{figure}

\section*{Appendix C: Navigation Visualization Analysis}
In Sec. IV-C4, we provide long-horizon runs with attached videos. In Fig.~\ref{fig:vid}, we present a representative frame. In addition to the adaptability we emphasize, we also note that the hyperparameters output under similar scene descriptions are highly consistent, demonstrating strong robustness.

\begin{figure}[htbp]
    \centering
    \includegraphics[width=1.0 \linewidth]{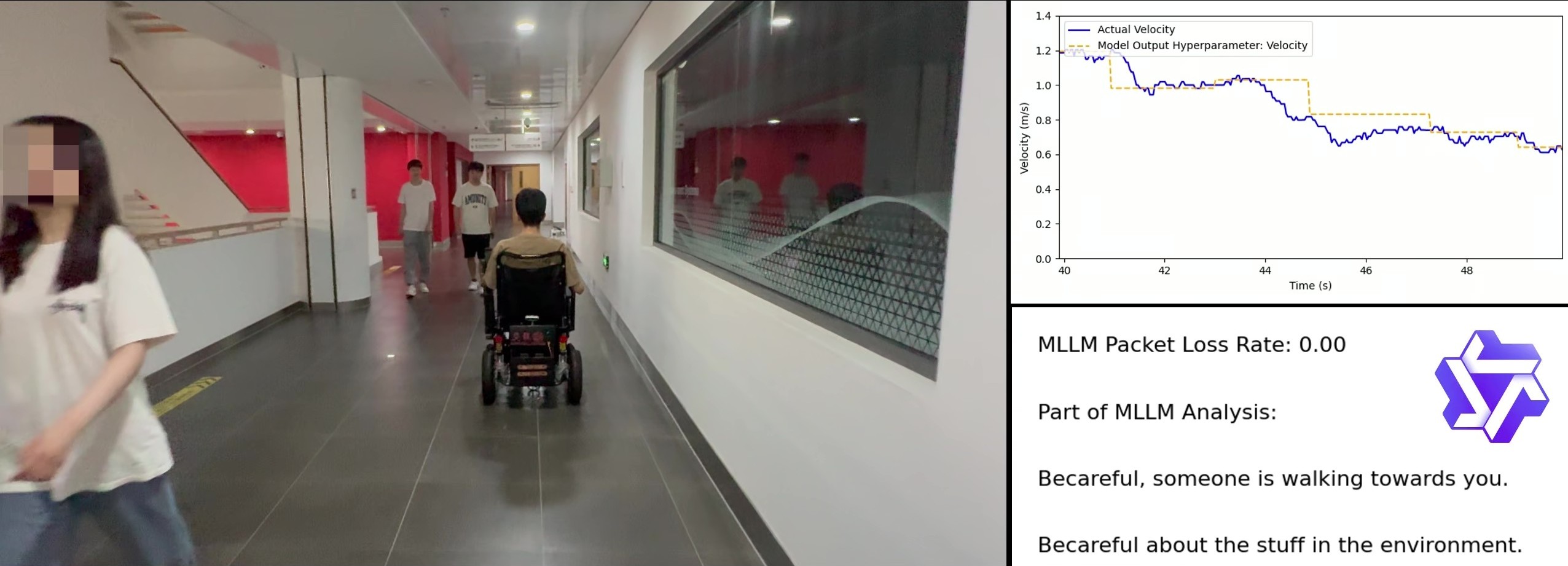}
    \caption{A sample frame from the demo. The video illustrates how MLLM-based analysis indirectly influences hyperparameter adjustments, and how the generative model enables more continuous and real-time tuning compared to human experts.}
    \label{fig:vid}
\end{figure}

}

% \clearpage

\bibliographystyle{IEEEtran}
\bibliography{main}

% \newpage

% \section{Biography Section}
% If you have an EPS/PDF photo (graphicx package needed), extra braces are
%  needed around the contents of the optional argument to biography to prevent
%  the LaTeX parser from getting confused when it sees the complicated
%  $\backslash${\tt{includegraphics}} command within an optional argument. (You can create
%  your own custom macro containing the $\backslash${\tt{includegraphics}} command to make things
%  simpler here.)
 
% \vspace{11pt}

% \bf{If you include a photo:}\vspace{-33pt}
% \begin{IEEEbiography}[{\includegraphics[width=1in,height=1.25in,clip,keepaspectratio]{fig1}}]{Michael Shell}
% Use $\backslash${\tt{begin\{IEEEbiography\}}} and then for the 1st argument use $\backslash${\tt{includegraphics}} to declare and link the author photo.
% Use the author name as the 3rd argument followed by the biography text.
% \end{IEEEbiography}

% \vspace{11pt}

% \bf{If you will not include a photo:}\vspace{-33pt}
% \begin{IEEEbiographynophoto}{John Doe}
% Use $\backslash${\tt{begin\{IEEEbiographynophoto\}}} and the author name as the argument followed by the biography text.
% \end{IEEEbiographynophoto}

\vfill

\end{document}